\documentclass[sigconf]{acmart}
\usepackage{algorithm}
\usepackage{algpseudocode}
\usepackage{booktabs}
\usepackage{array}
\usepackage{multirow}   
\usepackage{makecell}
\usepackage{graphicx}
\usepackage{balance}
\usepackage{flushend}
\usepackage{colortbl}   
\usepackage{xcolor}  
\usepackage{caption}
\AtBeginDocument{%
  }

\setcopyright{acmlicensed}
\copyrightyear{2018}
\acmYear{2018}
\acmDOI{XXXXXXX.XXXXXXX}
\setcopyright{rightsretained}
\copyrightyear{2026}
\acmYear{2026}
\acmDOI{10.1145/XXXXXXX.XXXXXXX} 
\acmConference[]{}{}{}
\acmISBN{978-1-4503-XXXX-X/26/03}

\acmISBN{978-1-4503-XXXX-X/2018/06}

\begin{document}

\title{Mind the Links: Cross-Layer Attention for Link Prediction in Multiplex Networks}
\settopmatter{printacmref=false}
\renewcommand\footnotetextcopyrightpermission[1]{}

\author{Devesh Sharma}
\authornote{This work was done while the author was at India Institute of Science Education Research Bhopal (IISERB).}
\email{deveshsawarni@gmail.com}
\affiliation{%
  \institution{IISERB}
  \country{}
}

\author{Aditya Kishore}
\email{adityak21@iiserb.ac.in}
\affiliation{%
  \institution{IISERB}
  \country{}
}

\author{Ayush Garg}
\email{ayushg24@iiserb.ac.in}
\affiliation{%
  \institution{IISERB}
  \country{}
}

\author{Debajyoti Mazumder}
\email{debajyoti22@iiserb.ac.in}
\affiliation{%
  \institution{IISERB}
  \country{}
}

\author{Debasis	Mohapatra}
\email{debasi.cse@pmec.ac.in}
\affiliation{%
  \institution{PMEC Berhampur}
  \country{}
}

\author{Jasabanta Patro}
\email{jpatro@iiserb.ac.in}
\affiliation{%
  \institution{IISERB}
  \country{}
}

\renewcommand{\shortauthors}{Sharma et al.}


\begin{abstract}
Multiplex graphs capture diverse relations among shared nodes. Most predictors either collapse layers or treat them independently. This loses crucial inter-layer dependencies and struggles with scalability. To overcome this, we frame multiplex link prediction as \emph{multi-view edge classification}. For each node pair, we construct a sequence of per-layer edge views and apply cross-layer self-attention to fuse evidence for the target layer. We present two models as instances of this framework: \textsc{\textbf{Trans-SLE}}, a lightweight transformer over static embeddings, and \textsc{\textbf{Trans-GAT}}, which combines layer-specific GAT encoders with transformer fusion. To ensure scalability and fairness, we introduce a Union--Set candidate pool and two leakage-free protocols: cross-layer and inductive subgraph generalization. Experiments on six public multiplex datasets show consistent macro-\(F_1\) gains over strong baselines (MELL, HOPLP-MUL, RMNE). Our approach is simple, scalable, and compatible with both precomputed embeddings and GNN encoders.
The code is available \href{https://github.com/DSaWarni/Mind-the-Links-Cross-Layer-Attention-for-Multiplex-Link-Prediction}{here.}
\end{abstract}



\maketitle

\vspace{-1mm}
\section{Introduction}

\begin{figure}[ht]
  \centering
  \includegraphics[width=0.7\columnwidth]{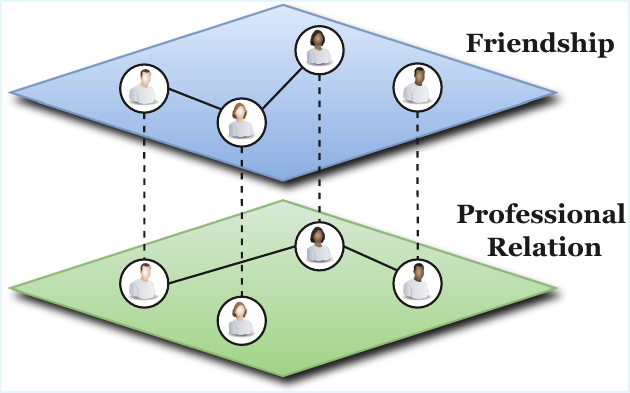}
  \caption{Example of a multiplex network.}
  \label{fig:Intro_1}  
  \Description{Two-layer diagram showing friendship (top) and professional relation (bottom). Dotted vertical links indicate the same node across layers.}
\end{figure}

Networks in online platforms, transportation, and biology rarely exist in a single dimension. The same entities interact in \emph{multiple ways}—friends exchange messages, colleagues coauthor papers, cities are linked by roads and flights. A \emph{multiplex} (multi-layer) network models this reality by placing the same node set across several layers, each layer encoding a distinct relation \cite{magnani2013combinatorial,kapferer,ckm,krackhardt1987cognitive}. In \autoref{fig:Intro_1}, we illustrate a simple example with ``friendship'' and ``professional'' layers: knowing that two users frequently collaborate can change how we interpret their sparse friendship signals, and vice-versa. The central question in this paper is: \emph{how can we predict a missing link on one layer while using evidence scattered across all layers?}

Link prediction, the task of inferring missing or future edges, has long been central to graph machine learning \cite{PhysRevResearch.2.042029,TANG2020105598,mishra2023hoplp,9394590,10132424,LUO2021106904}. Methods evolved from heuristics to embeddings like DeepWalk, LINE, and node2vec, and more recently, neural models like R-GCN and GraphGAN \cite{kumar2020link}. However, these methods typically flatten multiplex graphs or treat layers independently, ignoring crucial inter-layer dependencies and struggling with scalability and sparsity \cite{1313223}.
While multiplex networks model richer relationships \cite{10.1145/3012704}, they introduce new challenges: fusing cross-layer signals, handling imbalance, and managing the exponential growth of candidate edges. To address these challenges, our work introduces novel approaches for multiplex link prediction, framed as \emph{multi-view edge classification}. For each node pair, we form a sequence of per-layer edge views and use \emph{cross-layer self-attention} to integrate the most informative signals for a target layer. This enables us to model complex inter-layer dependencies effectively. We instantiate this idea in two models: \textsc{Trans-SLE} is a lightweight Transformer that operates on frozen node embeddings for scalability, while \textsc{Trans-GAT} augments this by adding per-layer GAT encoders to learn richer, layer-specific structure. To scale, we introduce a \emph{Union--Set} candidate pool: the union of observed edges across layers, which reduces computation while preserving meaningful pairs. Our approach is evaluated under leakage-free protocols with macro \(F_1\) scores on public benchmarks, showing consistent gains over strong baselines.
\\
\begin{figure*}[t!]
    \centering
    \includegraphics[width = \textwidth]{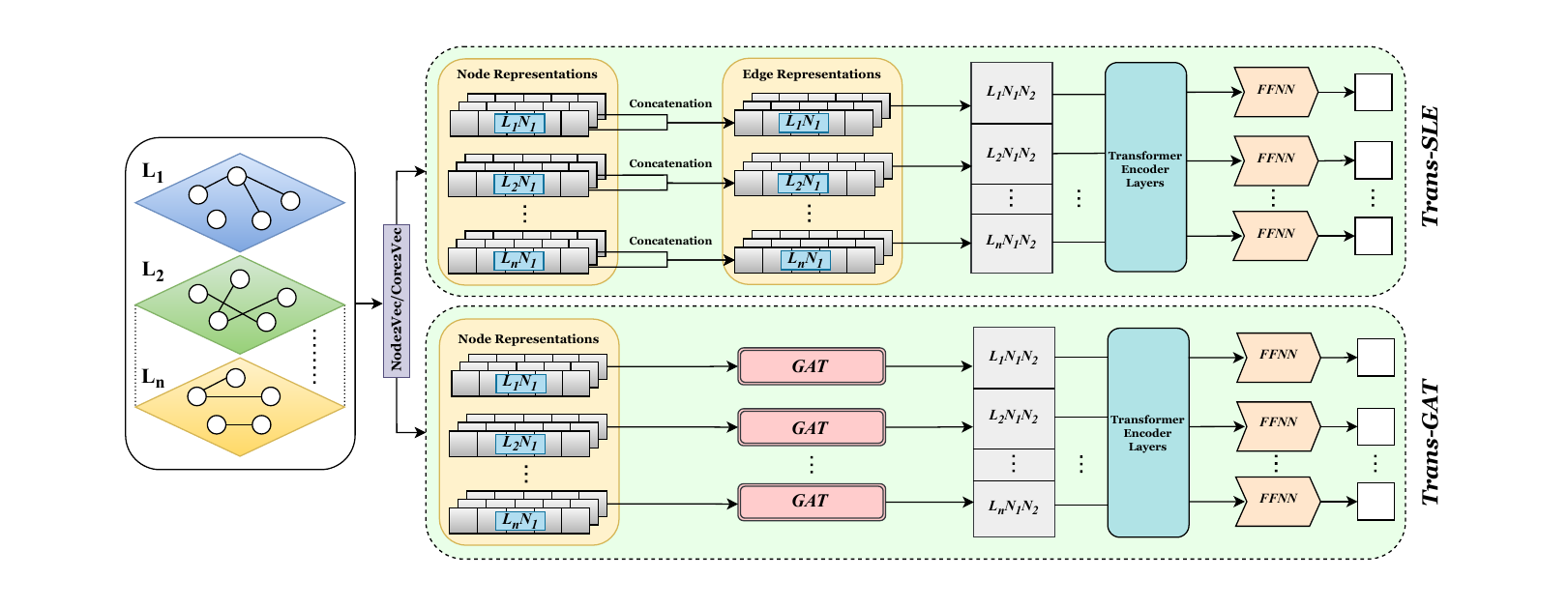}  
    \caption{ \textit{Trans-SLE} and \textit{Trans-GAT} architectures for link prediction in multiplex networks.}  
    \label{fig:main_arch}
    \Description{trans-GAT}
\end{figure*}

\noindent \textbf{Our contributions:}  
\begin{itemize}
  \item We frame multiplex link prediction as \emph{multi-view edge classification} with cross-layer attention.  
  \item We introduce a scalable \emph{Union--Set} candidate pool and leakage free evaluation protocols.
  \item We propose two transformer-based models: \textsc{Trans-SLE} (static embeddings) and \textsc{Trans-GAT} (end-to-end GNNs) beat the current state-of-the-art baselines.  
  
\end{itemize}

\section{Preliminaries}
\label{preliminaries}

We formalize a multiplex network as $\mathcal{G}=\{G^1,G^2,\ldots,G^l\}$, where each layer $G^i=(\mathcal{V}^i,\mathcal{E}^i)$ encodes a distinct type of relation among nodes $\mathcal{V}^i$ with edges $\mathcal{E}^i$. The total number of edges is $\sum_{i=1}^l |\mathcal{E}^i|$, and the maximum number of edges in a multiplex network with $N$ nodes is $l \cdot N(N-1)/2$. To reduce the computational cost, we define the \emph{union set} of edges as
\[
\mathcal{E}_{\text{union}} = \bigcup_{i=1}^{l} \mathcal{E}^i,
\]
which includes all edges observed across layers, assuming that node pairs without interaction in any layer can be ignored.
The task is to predict missing links in a target layer using information from the other layers. We adopt \emph{subgraph-level learning}, where models are trained on subgraphs $\{s^i \subseteq G^i\}$ from source layers and evaluated on unseen subgraphs $H' \subseteq H$ of a target graph $H$. The objective is to learn a function $f$ that predicts links in subgraphs and generalizes to new structures.
\section{Proposed Methods}
In this section, we present our frameworks for multiplex link prediction, illustrated in \autoref{fig:main_arch}, formulated as \emph{multi-view edge classification}. The key idea is to predict missing links by capturing how different layers jointly influence node relationships. For each node pair $(u,v)$, we form a \emph{multi-view edge sequence}, where each element encodes the pair’s embedding in a specific layer:
\[
Input_{(i,j)} = \{(u^m_i,v^m_j)\;|\;(u_i,v_j)\in U,\; m=1,\ldots,l\},
\]
with $U$ as the union set of candidate edges and $l$ the number of layers. A Transformer encoder with self-attention then integrates these layer-wise embeddings, emphasizing informative layers while suppressing irrelevant ones. The Union--Set reduces computational cost without discarding meaningful edges. We introduce two models \textsc{Trans-SLE} and \textsc{Trans-GAT}, that use cross-layer attention to capture multiplex dependencies while remaining efficient and generalizable (see Secs.~\ref{trans-sle_main} and~\ref{trans-gat-main}).

\subsection{Trans-SLE}
\label{trans-sle_main}
$\textsc{Trans-SLE}$, the first framework in our approach, is designed to classify candidate edges $(u,v)$ in multiplex networks. Its strength lies in capturing how different layers jointly influence the likelihood of a link. Drawing on the \textbf{Transformer architecture} \cite{vaswani2017attention}, we treat each edge as a sequence of layer-specific embeddings and apply self-attention to identify which layers provide the most informative cues for predicting a link in a designated target layer.  \\

\noindent \textbf{Input Representation:}  
For each candidate edge $(u,v)$, we collect its embeddings across all $'l'$ layers: $\{e^1_{uv}, e^2_{uv}, \ldots, e^l_{uv}\},$ where $e^m_{uv}$ denotes the embedding of edge $(u,v)$ in layer $m$. We derive edge embeddings for pairs $(u,v)$ using pre-trained methods, such as Node2Vec \cite{grover2016node2vec} and Core2Vec \cite{Sarkar2018Core2VecAC}. These embeddings are fixed during training. This keeps the model agnostic to the embedding strategy and preserves layer-specific context. We prepend a randomly initialized [CLS] token as a global reference for classification, giving the final input sequence as: $x_{uv} = [\text{[CLS]}, e^1_{uv}, e^2_{uv}, \ldots, e^l_{uv}] \in \mathbb{R}^{(1+l)\times d_{\text{model}}},$ with $d_{\text{model}}=2d_{\text{node}}$.\\

\noindent \textbf{Transformer Encoder:}  
The sequence $x_{uv}$ is processed by a single self-attention layer, which dynamically weights layers according to their relevance for predicting the target link. Formally, for each position $j$, self-attention produces contextualized embeddings as:
$
z_j = \sum_{k} \alpha_{jk} W_v e_k, 
\quad 
\alpha_{jk} = \text{softmax}\!\left(\frac{(W_q e_j)(W_k e_k)^\top}{\sqrt{d}}\right),
$
where $W_q$, $W_k$, and $W_v$ are learnable projections, and attention weights $\alpha_{jk}$ highlight informative layers while down-weighting irrelevant ones. Through self-attention, the [CLS] token attends to all layer tokens and aggregates them into a single representation; its final hidden state is a learned summary of the cross-layer context. We pass this summary to a feedforward classifier: $p_{uv} = \sigma(W z_{\text{[CLS]}} + b)$,
yielding the probability of an edge in the target layer. This design leverages self-attention to capture inter-layer dependencies, fusing complementary information while filtering noise. We believe this use of the [CLS] token provides a lightweight global summary, making \textsc{Trans-SLE} both efficient and effective for multiplex link prediction.\\

\noindent \textbf{Target-Driven Learning:}  
Although \textsc{Trans-SLE} can score all layers, prediction is restricted to a designated target. The [CLS] token directs attention toward this layer, combining complementary signals. Final probabilities are $p_{uv}=\sigma(W z_{\text{[CLS]}}+b)$, optimized with binary cross-entropy. Only Transformer and classifier weights are trained, keeping embeddings fixed for stability and emphasizing cross-layer reasoning. This compact design captures global dependencies and sets the stage for \textsc{Trans-GAT}.

\subsection{Trans-GAT}
\label{trans-gat-main} 
While \textsc{Trans-SLE} models cross-layer dependencies with a Transformer, it relies on static embeddings (e.g., Node2Vec, Core2Vec) that miss fine-grained structure. \textsc{Trans-GAT} addresses this by inserting Graph Attention Networks (GATs) \cite{velivckovic2017graph} at each layer to learn embeddings from local neighborhoods. These layer-specific representations are then fed into the Transformer, which fuses them via cross-layer attention. This design combines GATs for structural detail with the Transformer’s ability to integrate signals across layers, yielding richer and more adaptive edge representations.\\

\textbf{Layer-Specific GAT Encoders:} For each layer $m \in \{1,\dots,l\}$ of the multiplex network, we introduce a dedicated GAT module that refines node representations using neighborhood attention:
\begin{align*}
h_i^{(m)} = \sigma\!\left(\sum_{j \in \mathcal{N}_i^{(m)}} \alpha_{ij}^{(m)} W^{(m)} h_j\right)
\end{align*}
\begin{align*}
\quad
\alpha_{ij}^{(m)} = \frac{\exp(\textit{LeakyReLU}(a^\top[W^{(m)} h_i \,\|\, W^{(m)} h_j]))}{\sum_{k \in \mathcal{N}_i^{(m)}} \exp(\cdot)},  
\end{align*}
where $h_i^{(m)}$ is the embedding of node $i$ in layer $m$, $W^{(m)}$ is a learnable weight matrix, $\mathcal{N}_i^{(m)}$ is node $i$’s neighborhood on layer $m$, and $\alpha_{ij}^{(m)}$ are normalized attention coefficients over neighbors. For each node pair $(u,v)$, we form an edge embedding by concatenating their GAT outputs: $e^{(m)}_{uv} = [h_u^{(m)} \,\|\, h_v^{(m)}]$.  \\

\textbf{Cross-Layer Transformer Fusion: } The sequence $\{e^{(1)}_{uv}, \dots, e^{(l)}_{uv}\}$ constitutes the multi-view input for pair $(u,v)$. A Transformer encoder with self-attention aggregates this sequence, highlighting informative layers and down-weighting irrelevant ones. A randomly initialized [CLS] token is appended to act as a global reference for classification, and its output embedding is passed through a feedforward layer to predict link probability: $p_{uv} = \sigma(W_{\text{cls}} z_{\text{cls}} + b),$
where $z_{\text{cls}}$ is the [CLS] embedding after Transformer fusion.\\

\textbf{Target Layer Prediction: }To focus learning on a designated target layer $t$, we deactivate its GAT module and replace its embedding with the [CLS] token. This ensures that information from auxiliary layers flows into the target prediction, avoiding leakage from the true structure of the target layer while still benefiting from cross-layer signals.\\

\textbf{Training: }The model is trained with weighted binary cross-entropy over the Union--Set candidate pool. Unlike \textsc{Trans-SLE}, which freezes precomputed embeddings, \textsc{Trans-GAT} jointly optimizes GAT and Transformer parameters, allowing it to refine layer-specific embeddings and improve cross-layer reasoning. By coupling GAT-based layer encoders with Transformer-based fusion, \textsc{Trans-GAT} captures both local structural nuances within layers and global dependencies across them. This hybrid design yields more informative edge representations and enhances accuracy in multiplex link prediction.
\begin{table}[t!]
\centering
\resizebox{\columnwidth}{!}{%
\begin{tabular}{l|cccc}
\toprule[0.12em]
\textbf{Dataset} & \textbf{\#Layers} & \textbf{\#Nodes} & \textbf{\#Edges} & \textbf{Domains} \\
\midrule
CS-Aarhus \cite{magnani2013combinatorial}         & 5  & 61   & 620   & Social \\
Vickers-Chan-7thGraders \cite{vickers1981representing} & 3  & 29   & 740   & Education \\
C.ELEGANS \cite{chen2006wiring}   & 3  & 279   & 5863  & Biological \\
PIERRE-AUGER \cite{de2015identifying} & 16  & 514  & 7153  & Astrophysics \\
Rattus-Genetic \cite{de2015structural}            & 6  & 2640 & 4268  & Genetic \\
EU-Air Transportation \cite{cardillo2013emergence}            & 37  & 450 & 3588  & Transport \\
\bottomrule[0.12em]
\end{tabular}%
}
\caption{Basic statistics of the datasets.}
\label{Table:Stats}
\end{table}

\begin{table*}[t]
\centering
\resizebox{\textwidth}{!}{%
\begin{tabular}{l|
  >{\columncolor{gray!15}}c| >{\columncolor{gray!15}} c |c |c |c |
  >{\columncolor{gray!15}}c | >{\columncolor{gray!15}}c |c |c |c}
\toprule[0.12em]
& \multicolumn{5}{c|}{F1 Score} & \multicolumn{5}{c}{ROC-AUC} \\
\cmidrule(lr){2-6}\cmidrule(lr){7-11}
Dataset
& \cellcolor{white}Trans-SLE & \cellcolor{white}Trans-GAT & MeLL & HOPLP & RMNE
& \cellcolor{white}Trans-SLE & \cellcolor{white}Trans-GAT & MeLL & HOPLP & RMNE \\
\midrule[0.12em]
CS-Aarhus*        & \textbf{0.67} & \textbf{0.67} & 0.62 & 0.37 & 0.56
                  & \textbf{0.75} & \textbf{0.77} & 0.70 & 0.60 & 0.65 \\
C.ELEGANS*         & \textbf{0.72} & \textbf{0.63} & 0.59 & 0.38 & 0.55
                  & \textbf{0.80} & \textbf{0.68} & 0.62 & 0.53 & 0.57 \\
Rattus-Genetic*   & \textbf{0.72} & \textbf{0.64} & 0.54 & 0.36 & 0.48
                  & \textbf{0.89} & \textbf{0.84} & 0.62 & 0.24 & 0.65 \\
Vickers*          & \textbf{0.66} & \textbf{0.57} & 0.53 & 0.39 & 0.55
                  & \textbf{0.81} & \textbf{0.65} & 0.64 & 0.05 & 0.62 \\
PIERRE-AUGER*     & \textbf{0.83} & \textbf{0.87} & 0.53 & 0.17 & 0.59
                  & \textbf{0.97} & \textbf{0.98} & 0.60 & 0.12 & 0.74 \\
Eu-Air*           & \textbf{0.83} & \textbf{0.83} & 0.52 & 0.45 & 0.52
                  & \textbf{0.97} & \textbf{0.97} & 0.59 & 0.44 & 0.58 \\
\bottomrule[0.12em]
\end{tabular}
}%
\caption{Merged results on multiplex datasets. Left block shows layerwise-averaged over 3 random seeds. \textbf{F1}, right block shows layerwise-averaged over 3 random seeds. \textbf{ROC-AUC}. An asterisk (*) indicates that Trans-SLE/Trans-GAT outperforms all baselines for that dataset.}
\label{tab:merged_f1_rocauc}
\end{table*}

\section{Experiments}
In this section, we report the dataset details, the baseline methods, and the evaluation strategies considered in the study.
\vspace{-1mm}
\subsection{Datasets}
We consider six multiplex datasets spanning across social, education, business, healthcare, astrophysics, transportation, and biological domains. They vary in scale and layer count, enabling evaluation under diverse conditions. These benchmarks are standard in prior work \cite{zhang2022role, matsuno2018mell, mishra2023hoplp}. Dataset statistics are presented in Table~\ref{Table:Stats}.

\subsection{Baselines}
We consider three baselines for our study: \textbf{MeLL} \cite{matsuno2018mell}, \textbf{RMNE} \cite{zhang2022role}, and \textbf{HOPLP-MUL} \cite{mishra2023hoplp}. MeLL is a simple embedding layer factor model, which learns shared node embeddings plus a per-layer vector that modulates link scores. On the other hand, RMNE performs a role-aware and role-modified random walks (within-layer and cross-layer “role” hops) followed by Skip-gram produced role-preserving embeddings. Finally, HOPLP-MUL is a non-neural path aggregator that counts higher-order walks across layers with geometric decay and density weights, giving an interpretable multi-hop structural baseline. Our model depends on attention-based methods that are contrastive to the concept of layer-factorized embedding (MeLL), parameter-free path counting (HOPLP-MUL), and role-aware representation learning (RMNE). We regenerated all three baselines on the considered datasets, using their reported settings to ensure a faithful comparison.

\subsection{Experimental Setup}
We use a stratified split of 70/15/15 for train/validation/test sets. To report the performance of all of our models, we consider 3 random seeds (42, 18, 45) and report the average of these seeds for reproducibility fairness. To prevent leakage, we follow an inductive protocol that withholds nodes entirely, forcing the model to make predictions for unseen nodes at test time. Training uses class-weighted logistic loss with Adam optimizer using gradient clipping, and dropout in layers, with early stopping based on validation macro-F1; the decision threshold is selected on validation and kept fixed for test. We use Macro F1 as the primary evaluation metric as we are treating this problem as a classification task. Additionally, we also report the ROC-AUC score as considered in the baselines. 

\section{Results and Discussion}
\subsection{Results}
We report our experimental results in Table~\ref{tab:merged_f1_rocauc}, which shows consistent gains for our attention-based models. 
\begin{itemize}
    \item In terms of \textbf{Macro-F1}, \textsc{Trans-SLE} registers an improvement by ($\sim \textbf{8.00}\% \uparrow$) on \textit{CS-Aarhus} (0.67 vs 0.62), ($\sim \textbf{22.03}\% \uparrow$) on \textit{C.ELEGANS} (0.72 vs 0.59), ($\sim \textbf{33.33}\% \uparrow$) on \textit{Rattus-Genetic} (0.72 vs 0.54), ($\sim \textbf{20.00}\% \uparrow$) on \textit{Vickers} (0.66 vs 0.55), ($\sim \textbf{40.68}\% \uparrow$) on \textit{PIERRE-AUGER} (0.83 vs 0.59), and ($\sim \textbf{59.61}\% \uparrow$) on \textit{Eu-Air} (0.83 vs 0.52). Similarly,  \textsc{Trans-GAT} shows resembling trend by improvement of ($\sim \textbf{3.64}\% $) to ($\sim \textbf{59.61}\% $) on the same datasets, respectively.\\

    \item  For \textbf{ROC--AUC}, the improvements are similarly strong. \textsc{Trans-SLE} and \textsc{Trans-GAT} exceed the best baseline by\\ ($\sim \textbf{7.14 - 64.40}\% \uparrow$) and ($\sim \textbf{1.56 - 64.40}\% \uparrow$), respectively on all the considered datasets.\\

    \item \textsc{Trans-GAT} seems to have an upper edge when the network is more layered (e.g., \textit{PIERRE-AUGER}: 0.87 vs 0.83 F1; 0.98 vs 0.97 AUC) and matches \textsc{Trans-SLE} on \textit{CS-Aarhus} (0.67 F1; 0.77 vs 0.75 AUC). \textsc{Trans-SLE} on the other hand remains the best cost–effective choice on less layered graphs and still delivers large gains over baselines (e.g., \textit{Vickers}: +0.11 F1; \textit{Eu-Air}: +0.31 F1).\\

\end{itemize}

\subsection{Discussion}
\begin{itemize}
    \item Our results highlight the importance of cross-layer attention in multiplex link prediction. The weighting of layers allows our models to emphasize informative signals while suppressing noise. This is evident in datasets with weakly informative layers, where attention provides a consistent advantage.\\

    \item We observe a trade-off between Trans-SLE and Trans-GAT. Trans-SLE is lightweight, scalable, and competitive when embeddings already capture meaningful structure. Trans-GAT, on the other hand, learns richer representations and achieves higher accuracy on larger and denser graphs. This distinction suggests that users can select a model depending on whether efficiency or structural fidelity is the main priority.\\

    \item With an increasing number of layers and nodes, Trans-SLE scales linearly with input size. Trans-GAT is more expensive because of per-layer GATs, but remains tractable for medium-sized networks. These findings highlight that our methods are both reliable and deployable on real-world multiplex networks.\\

    \item Finally, our study underlines a larger point. Multiplex link prediction benefits from treating edges as multi-view sequences instead of merging or isolating layers. By modeling cross-layer dependencies explicitly, we obtain consistent gains over strong baselines. This insight opens the door for further exploration of attention-based fusion and other sequence modeling approaches in graph learning.\\
\end{itemize}

\section{Conclusion:}
We treat each potential edge as a sequence of layer-wise views and fuse them with cross-layer attention. Methods of this idea, Trans-SLE and Trans-GAT, consistently outperform strong baselines across varied domains, while offering a practical trade-off between efficiency (SLE) and structural fidelity (GAT).
The Union-Set helps in keeping the setting scalable. The formulation is modular: it works with precomputed embeddings or learnable encoders. 
We see this as a useful template for multiplex graphs: a small amount of attention, applied at the right level, goes a long way, and opens clear paths for future work on interpretable fusion, calibration, and cost-aware deployment in large multiplex systems.

\balance

\newpage
\bibliographystyle{ACM-Reference-Format}
\bibliography{software}

\end{document}